\documentclass{article}
\usepackage{spconf,amsmath,graphicx,hyperref}

\usepackage{amssymb}  % newly added by shusen ma
\usepackage{multirow}  % newly added by shusen ma
\usepackage{booktabs}  % newly added by shusen ma

\title{IBN: An Interpretable Bidirectional-Modeling Network for Multivariate Time Series Forecasting with Variable Missing}
\name{Shusen Ma$^{\dag}$\textsuperscript{\rm 1}, Tianhao Zhang$^{\dag}$\textsuperscript{\rm 1}, Qijiu Xia\textsuperscript{\rm 1} and Yun-Bo Zhao$^{*}$\textsuperscript{\rm 1} \thanks{This work was supported by the National Natural Science Foundation of China (No. 62173317).  $^*$Corresponding authors. $^{\dag}$Equal contribution.}}
\address{\textsuperscript{\rm 1}University of Science and Technology of China, Hefei, China}
\begin{document}
%\ninept
%
\maketitle
\begin{abstract}
Multivariate time series forecasting (MTSF) often faces challenges from missing variables, which hinder conventional spatial–temporal graph neural networks in modeling inter-variable correlations. While GinAR addresses variable missing using attention-based imputation and adaptive graph learning for the first time, it lacks interpretability and fails to capture more latent temporal patterns due to its simple recursive units (RUs). To overcome these limitations, we propose the \textbf{I}nterpretable \textbf{B}idirectional-modeling \textbf{N}etwork (\textbf{IBN}), integrating Uncertainty-Aware Interpolation (UAI) and Gaussian kernel-based Graph Convolution (GGCN). IBN estimates the uncertainty of reconstructed values using MC Dropout and applies an uncertainty-weighted strategy to mitigate high-risk reconstructions. GGCN explicitly models spatial correlations among variables, while a bidirectional RU enhances temporal dependency modeling. Extensive experiments show that IBN achieves state-of-the-art forecasting performance under various missing-rate scenarios, providing a more reliable and interpretable framework for MTSF with missing variables. Code is available at: https://github.com/zhangth1211/NICLab-IBN.
\end{abstract}


\section{Introduction}
\label{sec:intro}

%Multivariate time series forecasting (MTSF) has been widely applied in real-world scenarios, such as traffic flow prediction \cite{11099061} and weather forecasting \cite{angryk2020multivariate}. Existing MTSF approaches are mostly built upon end-to-end learning frameworks (Transformer-based \cite{10888746, 10552140}, CNN-based \cite{wu2023timesnet}, and MLP-based \cite{zeng2023transformers}) or representation learning frameworks \cite{ma2025c3rl}. While these methods can effectively capture temporal dependencies of variables, they are often insufficient in modeling the spatial correlations among variables. Recently, spatial-temporal graph neural networks (STGNNs) have integrated temporal and spatial modeling, achieving remarkable performance in MTSF \cite{chen2023higher}.

Time series forecasting (TSF) is widely used in real-world applications such as traffic flow \cite{11099061} and weather prediction \cite{angryk2020multivariate}. Existing TSF methods are mostly based on end-to-end learning frameworks (Transformer-based \cite{10552140}, CNN-based \cite{wu2023timesnet}, and MLP-based \cite{zeng2023transformers}) or representation learning \cite{ma2025c3rl}. While effective in capturing temporal dependencies, these methods often fall short in modeling spatial correlations among variables. Recently, spatial–temporal graph neural networks (STGNNs) have integrated temporal and spatial modeling, achieving strong performance in multivariate time series forecasting (MTSF) \cite{chen2023higher}.

%Spatial–temporal graph neural networks (STGNNs) typically rely on complete datasets to construct the topological graph among variables, thereby learning their spatial correlations. However, in real-world scenarios, data may be missing due to sensor or device failures. In particular, certain applications, such as air quality forecasting \cite{yu2024ginar, pachal2022sequence}, may even encounter variable-level missing data. In such cases, existing STGNNs \cite{10457027} struggle to accurately model the spatial correlations between observed and missing variables. To address this issue, GinAR \cite{yu2024ginar}—the first work to investigate multivariate time series forecasting (MTSF) under variable-missing conditions—introduces an interpolation attention mechanism and an adaptive graph convolution module to reconstruct missing variable values and to dynamically learn the inter-variable graph structure, respectively, while employing simple recursive units (RUs) to capture temporal dependencies.

STGNNs typically rely on complete datasets to construct variable graphs and learn spatial correlations. However, real-world data often contain missing values due to sensor failures. In some applications, such as air quality forecasting \cite{yu2024ginar, pachal2022sequence}, entire variables may be missing, posing challenges for conventional STGNNs \cite{10457027} in modeling spatial dependencies between observed and missing variables. To tackle this, GinAR \cite{yu2024ginar}, the first work to investigate variable missing, introduces an interpolation attention (IA) mechanism and an adaptive graph convolution module to reconstruct missing variables and dynamically learn inter-variable graphs, while using simple recursive units (RUs) for temporal modeling.

Although GinAR achieves state-of-the-art (SOTA) forecasting performance, it lacks interpretability in reconstructing missing variables and in constructing inter-variable graph structures, and its simple RU module is insufficient to capture latent temporal patterns. Specifically, 1) While GinAR leverages a learnable variable embedding matrix to model inter-variable correlations and uses an attention mechanism to impute missing values based on these correlations, it does not assess the reliability of the reconstructed results, where the uncertainty of the imputed values can compromise the stability of temporal modeling. 2) The adaptive graph in GinAR captures spatial correlations between variables solely through adaptive learning of the embedding matrix, without explicitly modeling the spatial relationships among variables. 3) The simple RU framework in GinAR considers only unidirectional temporal dependencies, failing to capture symmetric dependency patterns.

\begin{figure*}[ht]
	\centering
	\includegraphics[width=\textwidth]{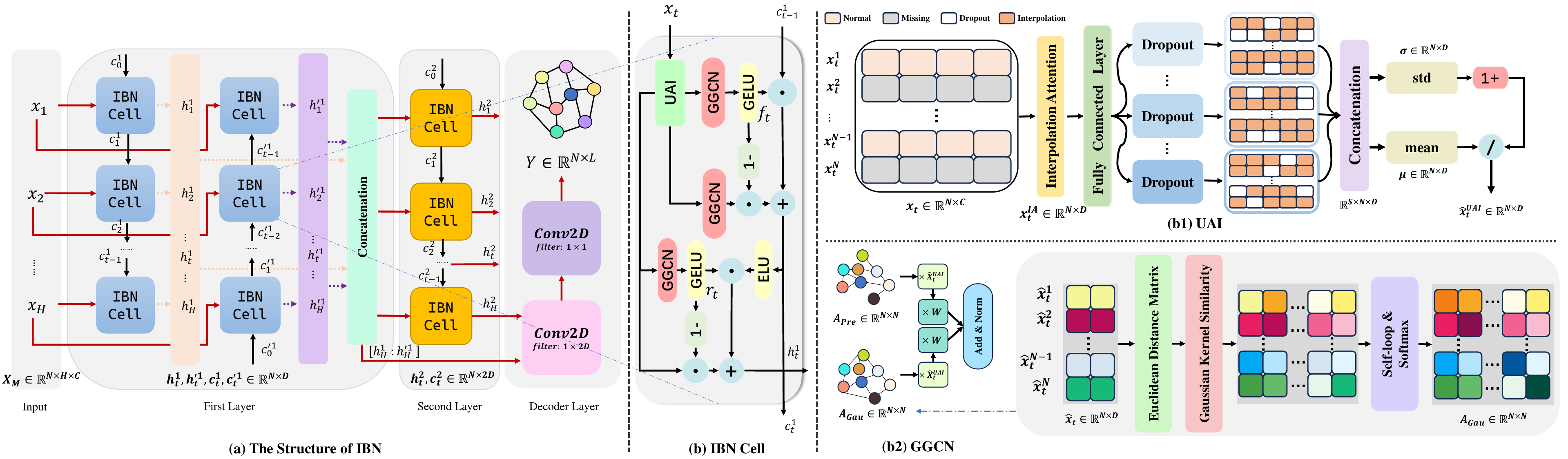}
	\caption{The structure of the IBN.}
	\label{IBN_structure}
\end{figure*}

To address the aforementioned challenges, we propose an Interpretable Bidirectional-modeling Network (IBN) for MTSF with variable missing, built upon Uncertainty-Aware Interpolation (UAI) and Gaussian kernel-based Graph Convolution (GGCN).
Technically, IBN first incorporates an uncertainty-aware mechanism during the reconstruction of missing variables: it leverages MC Dropout \cite{gal2016dropout} to estimate the distribution of reconstructed values and adopts an uncertainty-weighted strategy to attenuate the impact of high-risk reconstructions. Next, IBN employs GGCN to construct a more interpretable distance space, explicitly modeling the variations of spatial correlations among variables. Finally, a bidirectional RU (Bi-RU) architecture is designed to fully exploit temporal context information, thereby enhancing the temporal modeling capacity of the network. The main contributions of this work are as follows:
(1) We propose a more interpretable bidirectional forecasting framework, IBN, tailored for scenarios with variable missing patterns;
(2) By integrating UAI and GGCN, IBN improves both the reliability of reconstructed values for missing variables and the explicit modeling of spatial correlations among variables;
(3) Extensive experiments demonstrate that IBN consistently achieves SOTA forecasting performance across various missing rates on four public datasets.

\section{Background and Related Work}
\label{sec:bg}
Recent MTSF models primarily focus on capturing temporal dependencies within sequences \cite{zeng2023transformers}. While some studies have also considered inter-variable correlations, they typically model such relationships by learning the mutual influences among features intrinsic to each variable \cite{10552140}, overlooking the physical connections that exist in real-world scenarios. These physical connections are reflected in the inherent and observable topological structures among variables. In contrast, models based on STGNNs explicitly construct such physical topologies in the form of graphs and dynamically learn them, thereby enabling a more faithful modeling of the actual physical relationships among variables. For example, GTS \cite{shang2021discrete} leverages historical observations across all variables to infer the underlying discrete probabilistic graph structure, while MegaCRN \cite{jiang2023spatio} proposes a novel spatio-temporal meta-graph learning framework that improves traffic forecasting by addressing spatio-temporal heterogeneity and non-stationarity.

While the aforementioned studies are all conducted under the assumption of complete data, real-world datasets often suffer from missing values due to factors such as equipment failures or human errors. To address this issue, HSPGNN \cite{liang2024higher} integrates physical dynamics with dynamic graph learning to accurately impute missing values. BiTGraph \cite{chen2024biased} proposes a Biased Temporal Convolution Graph Network that explicitly captures missing patterns by incorporating biased temporal and spatial modules. LGnet \cite{tang2020joint} jointly models local and global temporal dynamics using a memory network and adversarial training for scenarios with missing data.
However, these works primarily focus on random missing values, whereas in practice, a more extreme form—variable missingness \cite{yu2024ginar, pachal2022sequence}—also occurs. In such cases, certain variables are entirely absent, making it challenging for existing approaches to effectively model the synergy between missing and observed variables. Although GinAR \cite{yu2024ginar} is the first to introduce a novel predictive framework for variable missingness, its reconstruction of missing variables and graph learning strategies still lack sufficient interpretability and robustness.

\section{Methodology}
\label{sec:method}

\subsection{ Problem Statement}
In MTSF, the evolution of each variable is influenced by both its own history and the dynamics of other variables. Such dependencies can be represented by a graph structure $G = (V, E)$, parameterized via an adjacency matrix $A \in \mathbb{R}^{N \times N}$. Here, $V$ denotes the nodes of the graph with $\| V \| = N$, $E$ represents the edges between nodes, and $N$ is the total number of variables.
MTSF aims to predict future values $Y \in \mathbb{R}^{N \times L}$ from historical observations $X \in \mathbb{R}^{N \times H \times C}$, where $L$ represents the forecasting length, $H$ denotes the historical window, and $C$ is the feature dimension. In the variable-missing scenario, certain variables exhibit complete historical absence, with their missing values replaced by zeros to obtain $X_M \in \mathbb{R}^{N \times H \times C}$. The objective is to learn a predictive mapping from $X_M$ to $Y$.

\subsection{Framework}
As shown in Fig. \ref{IBN_structure}(a), the IBN mainly consists of the first layer (Bi-RU), the second layer (simple RU), and the decoder layer. The input $X_M$ is processed by the Bi-RU to obtain hidden features with different temporal information: $Hid^{1} = \{ h_1^{1}, h_2^{1}, \dots, h_t^{1}, \dots, h_H^{1} \}$ and $Hid^{2} = \{ h_H^{'1}, h_{H-1}^{'1}, \dots, h_t^{'1}, \dots, h_1^{'1} \}$, where $\{ h_t^{1},h_t^{'1} \} \in \mathbb{R}^{N \times D}$, and $D$ denotes the dimension of the feature space. The number of RU cells equals $H$. $Hid^{1}$ and $Hid^{2}$ are concatenated along the last dimension to form the input of the unidirectional RU (Uni-RU), $x_t \in \mathbb{R}^{N \times 2D}$, whose hidden output is $h_t^{2} \in \mathbb{R}^{N \times 2D}$.
The decoder input is constructed by concatenating the last hidden outputs from the Bi-RU, $[ h_H^{1}, h_H^{'1} ] \in \mathbb{R}^{N \times 2D}$, with $h_H^{2} \in \mathbb{R}^{N \times 2D}$. The decoder contains two 2D convolutional layers, with kernel sizes of $1 \times 2D$ and $1 \times 1$, respectively, and the number of kernels in both layers is $L$.
The structure of the IBN Cell is illustrated in Fig. \ref{IBN_structure}(b): the UAI module is adopted as the input reconstructor of the simple RU, and the internal fully connected layer is replaced with the GGCN module to capture spatial correlations among variables. The formulation of the IBN Cell is as follows:
\begin{equation}
    f_t = \text{GeLU} \left( \text{GGCN}^{f_t}\left( \text{UAI}\left( x_t \right) \right) \right),
\end{equation}
\begin{equation}
    r_t = \text{GeLU} \left( \text{GGCN}^{r_t}\left(\text{UAI}\left( x_t \right) \right) \right),
\end{equation}
\begin{equation}
    c_t = f_t \odot c_{t-1} + \left( 1 - f_t \right) \odot \left( \text{GGCN}^{c_t}\left( \text{UAI}\left( x_t \right) \right) \right),
\end{equation}
\begin{equation}
    h_t = \left(1 - r_t \right) \odot \text{UAI} \left(x_t \right) + r_t \odot \text{ELU} \left( c_t \right) .
\end{equation}
Among them, $f_t$ denotes the forget gate, $r_t$ denotes the reset gate, $c_t$ denotes the cell state, and $h_t$ denotes the hidden state. $ \text{GeLU}(\cdot) $ and $ \text{ELU}(\cdot) $ represent activation functions. 
The following Section~\ref{UAI} and Section~\ref{GGCN} will introduce the structural principles of UAI and GGCN, respectively.

\subsubsection{UAI} \label{UAI}
Let $x_t^{IA} = \text{IA}\left( x_t \right) \in \mathbb{R}^{N \times D}$ denote the IA-completed input, where $\text{IA}(\cdot)$ is the Interpolation Attention \cite{yu2024ginar}.  
We then apply a linear transformation parameterized by $W \in \mathbb{R}^{D \times D}$ and $b \in \mathbb{R}^{D}$, followed by a dropout operator $D_p(\cdot)$ with drop rate $p$ (set to 0.1 in this paper). For the $s$-th stochastic forward pass, we have:
\begin{equation}
    \mathbf{m}^{(s)} = D_p\left( x_t^{\text{IA}} \cdot W + b \right), \quad s = 1, 2, \dots, S,
\end{equation}
where $S$ is the total number of Monte Carlo samples \cite{gal2016dropout} and is set to 10 in this paper.
Stacking all sampled hidden representations yields:
\begin{equation}
    \mathcal{M} = \{\mathbf{m}^{(1)}, \mathbf{m}^{(2)}, \dots, \mathbf{m}^{(S)} \} 
    \in \mathbb{R}^{S \times N \times D}.
\end{equation}
The predictive mean $\mu$ and standard deviation $\sigma$ are computed as: $ \mu = \frac{1}{S} \sum_{s=1}^S \mathbf{m}^{(s)}, 
    \quad
    \sigma = \sqrt{\frac{1}{S} \sum_{s=1}^S \left( \mathbf{m}^{(s)} - \mu \right)^2 } . $
%\begin{equation}
%    \mu = \frac{1}{S} \sum_{s=1}^S \mathbf{m}^{(s)}, 
%    \quad
%    \sigma = \sqrt{\frac{1}{S} \sum_{s=1}^S \left( \mathbf{m}^{(s)} - \mu \right)^2 } .
%\end{equation}
The UAI outputs the pair $(\mu, \sigma)$ as the prediction value and its uncertainty estimate. Finally, UAI uses the following formula to obtain the final values:
\begin{equation}
    {\hat{x}}_t^{\text{UAI}} = \frac{\mu}{1+\sigma}, 
\end{equation}
which means that the greater the uncertainty, the lower the model's attention weight assigned to the imputation result.

\subsubsection{GGCN} \label{GGCN}
As shown in Fig.~\ref{IBN_structure}(b2), GGCN is built upon the predefined graph $A_{Pre}$ and the dynamic graph $A_{Gau}$ constructed using a Gaussian kernel. The predefined graph $A_{Pre}$ is constructed based on known information (i.e., the geographical distances between nodes), which captures the direct connections among different nodes. However, relying solely on distance information between nodes is insufficient to fully capture the spatial dependencies embedded in the node features themselves. Therefore, it is necessary to construct a learnable dynamic graph based on node features to further explore the underlying spatial information. Unlike GinAR, which directly learns an adaptive graph through a variable-embedded matrix (leading to poor interpretability and instability), GGCN first computes the spatial distance between the features of nodes $i$ and $j$ using the Euclidean distance matrix $D_{i,j} = \left \| \hat{x}_t^{i} - \hat{x}_t^{j} \right \|_2, \hat{x}_t^{i} \in \mathbb{R}^{D}$. Since Euclidean distances can be very large (especially in high-dimensional feature spaces), directly applying the distance or its negative value to softmax may cause gradient explosion or vanishing. To address this, GGCN adopts a Gaussian kernel function to compress possible extreme values into the range $[0, 1]$, thereby stabilizing gradients during training and facilitating model convergence, as formulated below:
$ A_{i,j} = \exp\left( - \frac{ \left( D_{i,j} \right)^2 }{ 2\gamma } \right) $ ,
where $\gamma$ denotes the embedding size. Finally, the matrix $A$ is processed with the self-loop and softmax to obtain the adjacency matrix $A_{Gau} \in \mathbb{R}^{N \times N}$, which reflects the spatial correlation variations among nodes, as shown in the following formula:
\begin{equation}
    A_{Gau_{i,j}} = \frac{\exp\left(\tilde{A}_{i,j}\right)}{\sum_{k=1}^{N} \exp\left(\tilde{A}_{i,k}\right)}, \quad \tilde{A} = A + I_N .
\end{equation}
After obtaining $A_{Pre}$ and $A_{Gau}$, we assign them learnable weight matrices respectively and then combine them for joint training. This allows the model not only to perceive the static spatial dependencies between nodes, but also to capture the dynamic spatial relationships among the features contained in the nodes. The formulation is as follows:
\begin{equation}
    x_t^{\text{GGCN}} = \text{Norm} \left( \left( A_{Pre} \cdot {\hat{x}}_t^{\text{UAI}} \right) \cdot W + \left( A_{Gau} \cdot {\hat{x}}_t^{\text{UAI}} \right) \cdot W \right) ,
\end{equation}
where ${\hat{x}}_t^{\text{UAI}} \in \mathbb{R}^{N \times D}$ denotes the output of the UAI module.

\section{Experiments}
\label{sec:exp}
\textit{\textbf{Experiment setup}}. We use four public datasets (METR-LA, PEMS-BAY, PEMS04, and PEMS08 \cite{yu2024ginar}) as our benchmarks. Six recent models—MegaCRN \cite{jiang2023spatio}, LGnet \cite{tang2020joint}, GC-VRNN \cite{xu2023uncovering}, TriD-MAE \cite{zhang2023trid}, BiTGraph \cite{chen2024biased}, and GinAR \cite{yu2024ginar}—are selected as the baseline models for comparison. Three evaluation metrics (RMSE, MAPE, and MAE) are used to comprehensively assess the prediction performance of different baselines and IBN. We randomly mask variables at rates of 25\%, 50\%, and 75\% to evaluate the model’s generalization ability \cite{yu2024ginar}. The hyperparameter settings of IBN for different tasks can be found in our publicly available code.

\begin{table}[ht]
\centering
\caption{Performance comparison results of all baselines and the proposed model on all datasets. The best results are shown in \textbf{bold}, and the second best are \underline{underlined}.}
\label{full_results}
\renewcommand{\arraystretch}{1.2} % <--- 添加此行以缩小行距
\setlength{\tabcolsep}{3.5pt} % 列间距
%\footnotesize        % 比 \scriptsize 更合适阅读
\tiny
%\begin{adjustbox}{width=\textwidth}
\begin{tabular}{ccccccccccc}
\toprule
\multirow{2}{*}{\textbf{Datasets}} & \multirow{2}{*}{\textbf{Methods}} & \multicolumn{3}{c}{\textbf{Missing rate 25\%}} & \multicolumn{3}{c}{\textbf{Missing rate 50\%}} & \multicolumn{3}{c}{\textbf{Missing rate 75\%}}  \\
\cmidrule(lr){3-5} \cmidrule(lr){6-8} \cmidrule(lr){9-11} 
 & & RMSE & MAPE & MAE & RMSE & MAPE & MAE & RMSE & MAPE & MAE  \\
\midrule

\multirow{7}{*}{\textbf{METR-LA}} 
& MegaCRN & 7.43 & 10.47 & 3.81 & 7.87 & 11.02 & 3.94 & 8.28 & 12.13 & 4.24 \\
& LGnet & 7.52 & 10.97 & 3.95 & 8.03 & 11.83 & 4.17 & 8.52 & 13.09 & 4.42 \\
& GC-VRNN & 7.04 & 10.51 & 3.68 & 7.73 & 10.98 & 3.87 & 8.19 & 11.71 & 4.17  \\
& TriD-MAE & 7.15 & 10.37 & 3.64 & 7.58 & 11.07 & 3.79 & 7.92 & 11.13 & 3.92  \\
& BiTGraph & \underline{6.74} & 10.25 & 3.61 & 7.32 & 10.79 & 3.69 & 7.63 & 11.04 & 3.74  \\
& GinAR & \textbf{6.55} & 10.12 & 3.56 & 7.14 & 10.42 & \textbf{3.61} & 7.39 & \textbf{10.71} & \textbf{3.70}  \\
& \textbf{IBN} & 6.84 & \textbf{9.78} & \textbf{3.53} & \textbf{7.09} & \textbf{10.30} & \textbf{3.61} & \textbf{7.23} & \underline{10.98} & \underline{3.72}  \\
\midrule

\multirow{7}{*}{\textbf{PEMS-BAY}} 
& MegaCRN & 5.93 & 7.03 & 2.85 & 7.36 & 7.75 & 3.02 & 7.75 & 8.77 & 3.35  \\
& LGnet & 6.02 & 7.19 & 2.88 & 6.74 & 8.15 & 3.14 & 8.08 & 9.18 & 3.43  \\
& GC-VRNN & 4.93 & 5.37 & 2.39 & 5.34 & 5.86 & 2.64 & 6.08 & 6.94 & 2.87  \\
& TriD-MAE & 5.17 & 5.48 & 2.46 & 5.53 & 6.12 & 2.69 & 5.97 & 6.85 & 2.79  \\
& BiTGraph & \underline{4.52} & 5.16 & \underline{2.17} & 5.06 & 6.07 & 2.44 & 5.79 & 6.68 & 2.61  \\
& GinAR & \textbf{4.34} & \textbf{4.90} & \textbf{2.10} & \underline{4.78} & \underline{5.88} & \underline{2.35} & \underline{5.48} & \underline{6.17} & \underline{2.54}  \\
& \textbf{IBN} & 4.53 & \underline{5.03} & 2.22 & \textbf{4.69} & \textbf{5.32} & \textbf{2.27} & \textbf{5.34} & \textbf{6.02} & \textbf{2.53}  \\
\midrule

\multirow{7}{*}{\textbf{PEMS04}} 
    & MegaCRN & 42.22 & 20.18 & 28.26 & 47.07 & 21.29 & 31.48 & 47.95 & 22.03 & 33.58  \\
    & LGnet & 42.64 & 18.42 & 26.53 & 46.39 & 21.30 & 30.81 & 52.05 & 24.83 & 33.94  \\
    & GC-VRNN & 39.75 & 16.82 & 23.57 & 41.34 & 17.83 & 26.43 & 43.82 & 18.67 & 27.72  \\
    & TriD-MAE & 39.83 & 16.98 & 24.15 & 40.65 & 17.52 & 25.89 & 41.90 & 18.04 & 26.95  \\
    & BiTGraph & 38.94 & 16.75 & 23.01 & 40.03 & 17.34 & 24.15 & 41.69 & 17.92 & 26.33  \\
    & GinAR & \underline{38.22} & \underline{16.45} & \underline{22.52} & \underline{39.02} & \underline{17.04} & \underline{23.78} & \underline{41.53} & \underline{17.58} & \underline{25.98}  \\
    & \textbf{IBN} & \textbf{36.51} & \textbf{16.17} & \textbf{22.51} & \textbf{36.54} & \textbf{16.89} & \textbf{22.66} & \textbf{37.26} & \textbf{17.22} & \textbf{23.26}  \\
    \midrule
    
    \multirow{7}{*}{\textbf{PEMS08}} 
    & MegaCRN & 39.29 & 17.42 & 26.04 & 43.43 & 21.30 & 30.68 & 48.37 & 21.34 & 32.23  \\
    & LGnet & 37.54 & 22.18 & 26.51 & 47.23 & 21.91 & 32.04 & 50.38 & 21.43 & 33.65  \\
    & GC-VRNN & 35.17 & 14.69 & 23.25 & 36.40 & 15.85 & 24.27 & 39.67 & 16.06 & 26.32  \\
    & TriD-MAE & 33.15 & 14.25 & 21.53 & 35.95 & 15.32 & 23.18 & 37.64 & 15.58 & 24.89  \\
    & BiTGraph & 31.89 & 14.05 & 20.65 & \underline{35.06} & 14.62 & 22.44 & 36.98 & 15.04 & 23.38  \\
    & GinAR & \underline{31.34} & \underline{13.76} & \underline{20.41} & \textbf{34.53} & \underline{14.21} & \underline{22.01} & \textbf{36.04} & \underline{14.77} & \underline{23.10}  \\
    & \textbf{IBN} & \textbf{29.84} & \textbf{13.15} & \textbf{19.40} & 37.35 & \textbf{13.79} & \textbf{21.86} & \underline{36.14} & \textbf{14.59} & \textbf{22.07}  \\
    \bottomrule

\end{tabular}
%\end{adjustbox}
\end{table}

\begin{table}[ht]
\centering
\caption{Ablation study of IBN components on diverse datasets under the 50\% variable missing.}
\label{ablation_results}
\renewcommand{\arraystretch}{1.2}   % 稍微压缩行高
\setlength{\tabcolsep}{2.5pt}        % 缩小列间距，不要太小
%\footnotesize        % 比 \scriptsize 更合适阅读
\tiny
\begin{tabular}{cccccccccccccc}
\toprule
\multirow{2}{*}{\textbf{Datasets}} & \multicolumn{3}{c}{\textbf{IBN}} & \multicolumn{3}{c}{UAI $\rightarrow$ IA} & \multicolumn{3}{c}{GGCN $\rightarrow$ AGCN} & \multicolumn{3}{c}{Bi-RU $\rightarrow$Uni-RU} \\
\cmidrule(lr){2-4} \cmidrule(lr){5-7} \cmidrule(lr){8-10} \cmidrule(lr){11-13}
& RMSE & MAPE & MAE & RMSE & MAPE & MAE & RMSE & MAPE & MAE & RMSE & MAPE & MAE \\
\midrule

\multirow{1}{*}{\textbf{METR-LA}} 
& \textbf{7.09} & \textbf{10.30} & \underline{3.61} & 7.45 & 10.69 & 3.70 & \underline{7.12} & \underline{10.46} & \textbf{3.58} & 7.20 & 10.60 & 3.69 \\
\midrule

\multirow{1}{*}{\textbf{PEMS-BAY}} 
& \textbf{4.69} & \textbf{5.32} & \textbf{2.27} & 4.94 & 5.48 & 2.34 & \underline{4.79} & \underline{5.34} & \underline{2.32} & 4.88 & 5.58 & 2.37 \\
\midrule

\multirow{1}{*}{\textbf{PEMS04}} 
& \textbf{36.54} & \underline{16.89} & \textbf{22.66} & 37.14 & 17.09 & 23.01 & \underline{36.71} & 17.01 & \underline{22.88} & 37.06 & \textbf{16.55} & 23.05 \\
    \midrule
    
    \multirow{1}{*}{\textbf{PEMS08}} 
& 37.35 & \textbf{13.79} & 21.86 & \underline{36.72} & 14.92 & \underline{21.85} & 38.50 & \underline{14.51} & 22.47 & \textbf{36.12} & 14.59 & \textbf{21.59} \\
    \bottomrule

\end{tabular}
\end{table}

\textit{\textbf{Main results}}.
The results in Table~\ref{full_results} demonstrate that IBN achieves overall superior performance compared to GinAR. Specifically, IBN attains average MAE improvements of \textbf{1.89\%} (12.15 $\rightarrow$ 11.92), \textbf{2.63}\% (12.94 $\rightarrow$ 12.60), and \textbf{6.72}\% (13.83 $\rightarrow$ 12.90) over GinAR on 25\%, 50\%, and 75\% missing rates, respectively, where the average MAE refers to the mean MAE across four different datasets. This significant performance gain can be largely attributed to IBN’s uncertainty-aware reconstruction and its stronger modeling capacity for spatiotemporal dependencies.

\textit{\textbf{Ablation study}}.
To verify the effectiveness of the components in IBN, we conducted the following three groups of ablation experiments: replacing UAI with IA \cite{yu2024ginar}, replacing GGCN with AGCN \cite{yu2024ginar}, and replacing Bi-RU with a unidirectional RU \cite{yu2024ginar}. The results in Table~\ref{ablation_results} show that, compared with the core components in GinAR, the corresponding improved modules proposed in this work achieve better predictive performance while offering stronger interpretability and robustness, thereby demonstrating the rationality of the IBN component design.

\begin{figure}[ht]
	\centering
	\includegraphics[width=0.48\textwidth]{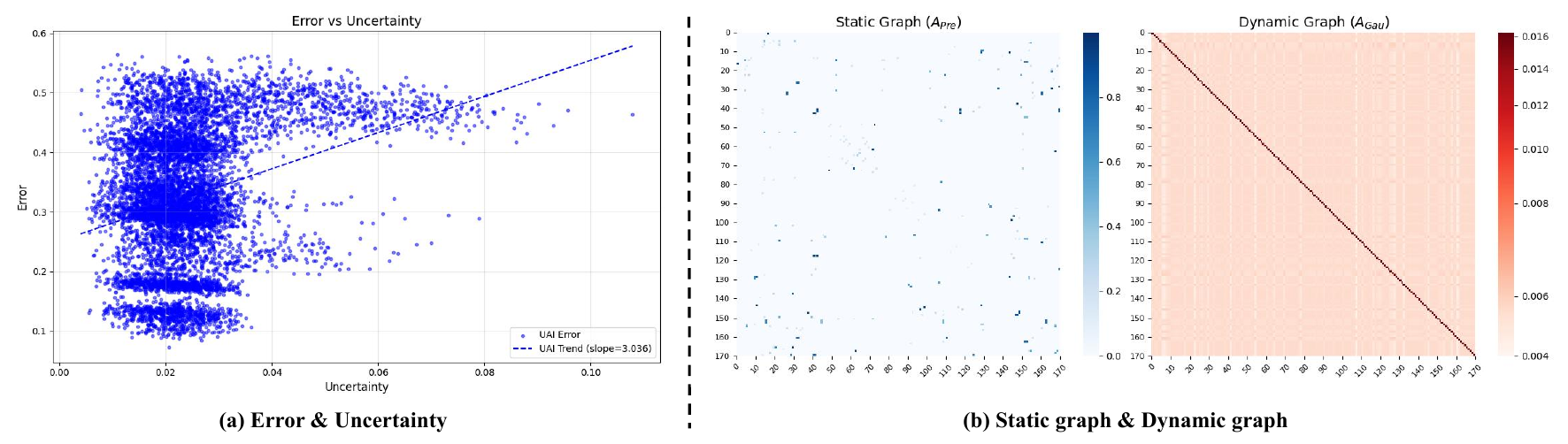}
	\caption{Interpretability analysis of the UAI and GGCN.}
	\label{interpretability_analysis}
\end{figure}

\textit{\textbf{Interpretability analysis}}.
To validate the interpretability of IBN, we visualized the error–uncertainty relationship of UAI and the inter-node attention weights learned by GGCN, as shown in the Fig.~\ref{interpretability_analysis}. Specifically, Fig.~\ref{interpretability_analysis}(a) illustrates that as the reconstruction error of UAI increases, the model’s estimated uncertainty for the corresponding prediction also rises, thereby reducing the attention weight assigned to that reconstructed value in the overall prediction. This demonstrates the model’s adaptive regulation capability in uncertainty modeling. Fig.~\ref{interpretability_analysis}(b) presents the inter-node attention weights learned by GGCN, where the left subfigure depicts stable static spatial topologies among nodes, while the right subfigure captures the dynamic spatial dependencies among node features over time. Together, the two subfigures demonstrate that GGCN can effectively capture and distinguish multi-level spatial structures, thereby enhancing the model’s representational capacity and interpretability.

\section{Conclusion}
\label{sec:conclusion}
We propose a novel model, IBN, to address the prevalent issue of variable missingness in multivariate time series forecasting tasks. Compared with existing approaches, IBN demonstrates significant advantages in both predictive performance and interpretability. Specifically, IBN incorporates the UAI mechanism, enabling the model to effectively evaluate the reliability of interpolation results and thereby enhance their credibility. In addition, IBN introduces a new graph neural network structure, GGCN, constructed on Gaussian kernels and node features, which not only strengthens model stability but also further improves interpretability. Moreover, IBN adopts a bidirectional recursive unit to build the overall framework, effectively enhancing the capability of capturing and modeling temporal features. Experimental results on four public datasets under varying missing rates show that IBN achieves SOTA performance in forecasting accuracy.

\bibliographystyle{IEEEbib}
\bibliography{refs}

\end{document}